\documentclass{styles/svproc}
\usepackage{url}
\usepackage{bm}
\usepackage{tikz}
\usepackage[caption=false, font=footnotesize]{subfig}
\usepackage{amssymb}
\usepackage{algorithm}
\usepackage{algorithmic}
\usepackage{hyperref}
\usepackage{marvosym}

\begin{document}
\mainmatter              
\title{Time-Optimized Trajectory Planning for Non-Prehensile Object Transportation in 3D}
\titlerunning{ Non-Prehensile Object Transportation}  
%
\author{Lingyun Chen\inst{1, 2}\textsuperscript{(\Letter)} \and Haoyu Yu \inst{1} \and Abdeldjallil Naceri \inst{1} \and Abdalla Swikir \inst{1, 2} \and Sami Haddadin \inst{1, 2}}
\authorrunning{Lingyun Chen et al.} 
%
\tocauthor{Lingyun Chen, Haoyu Yu, Abdeldjallil Naceri, Abdalla Swikir and Sami Haddadin}
\institute{Munich Institute of Robotics and Machine Intelligence (MIRMI), Technical University of Munich \\
\and Centre for Tactile Internet with Human-in-the-Loop (CeTI) \\
\email{lingyun.chen@tum.de}
}

\maketitle              

\vspace{-0.8cm}
\begin{abstract}
Non-prehensile object transportation offers a way to enhance robotic performance in object manipulation tasks, especially with unstable objects. Effective trajectory planning requires simultaneous consideration of robot motion constraints and object stability. Here, we introduce a physical model for object stability and propose a novel trajectory planning approach for non-prehensile transportation along arbitrary straight lines in 3D space. Validation with a 7-DoF Franka Panda robot confirms improved transportation speed via tray rotation integration while ensuring object stability and robot motion constraints.

\keywords{Trajectory planning, non-prehensile object transportation}
\end{abstract}

\vspace{-0.8cm}
\section{Introduction}

With the advancement of robotic technology, robots are finding increasingly widespread applications in industrial production\cite{hentout2019human}. In the industrial production process, the transportation of objects is undeniably crucial. Transporting objects through non-prehensile manipulation, compared to grasping objects, offers several advantages, including simpler end-effector design, the ability to transport a wider range of objects, and improved transportation efficiency\cite{mason1993dynamic}. 
\par
\begin{figure}[ht]
    \centering
    \vspace{-15pt}
    \includegraphics[width=0.50\textwidth]{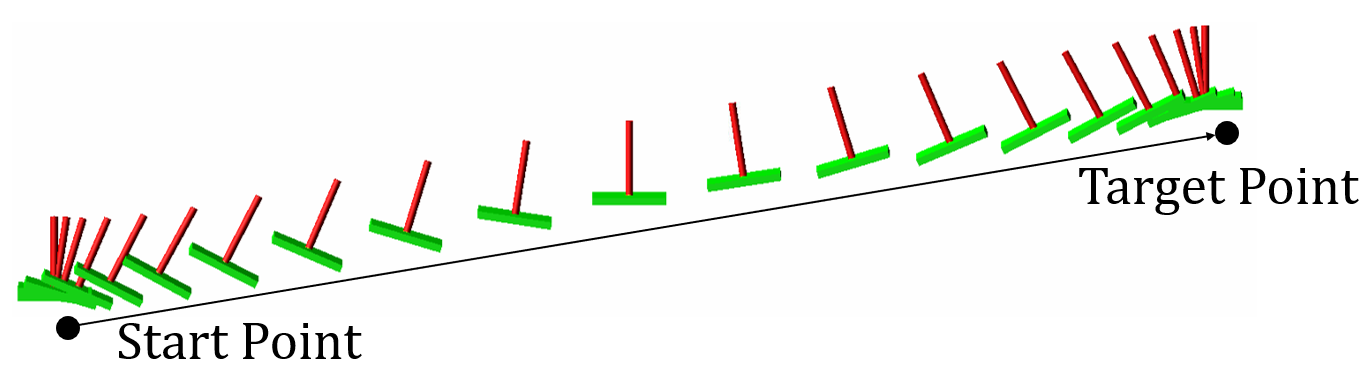}
    \vspace{-15pt}
    \caption{Illustration of object and tray motion.}
    \label{fig:illustration}
\end{figure}

Transporting objects using non-prehensile manipulation requires a low center of gravity, a large base area, and minimizing acceleration during the transportation process for stability. To address the transportation challenges of unstable objects, in \cite{acharya2020nonprehensile}, researchers explored the limiting conditions for the maximum acceleration during the transportation of unstable objects on a tray-like end-effector. They achieved planar transportation through trajectory planning based on an S-curve. However, their work does not incorporate the rotation of the tray. Introducing tray rotation at different stages of transportation could evidently enhance the time efficiency of the process. Relevant studies include those in \cite{tsuji2015dynamic}, which examine the contact model between the object and the tray, and \cite{selvaggio2023non}, which proposes a Model Predictive Control-based approach to track predefined transportation trajectories. The latter ensures non-sliding transportation of the object by considering the friction cone between the object and the tray as a constraint. Both papers discuss the rotation of the tray and the physical model of contact between the object and the tray. However, while similar studies introduce tray rotation, the majority focus their control objectives on enhancing tracking performance for given trajectories. In contrast, our work approaches the problem from a trajectory planning perspective, aiming to improve transportation speed by incorporating the rotational motion of the tray. The motion of the tray and the object in the desired trajectory is illustrated in Fig. \ref{fig:illustration}.
\vspace{-35pt}
\section{Method}
In subsections \ref{IIIA} and \ref{IIIB}, we discuss the physical model between the object and the tray and present the novel planning method for computing the transport trajectory, respectively.
\vspace{-10pt}
\subsection{Physical Modelling}\label{IIIA}
In our setting, it is presumed that both the object and the tray are rigid bodies. The object under consideration possesses a slender, uniform cylindrical morphology, and in the analysis, the effects of air resistance on its motion are disregarded. Furthermore, it is assumed that the object exhibits characteristics such as a high center of gravity and a minimal base area, consequently resulting in compromised stability. Under conditions of acceleration, the object is anticipated to undergo tipping prior to the onset of sliding between its base and the tray.
\par
\begin{figure}[ht!]
    \centering
    \vspace{-25pt}
    \subfloat[Contact model \label{fig:phy}]{    
    \begin{tikzpicture}[scale = 0.25]
        \draw (0,-0.5)--(14,-0.5);
        \draw (14,-0.5)--(0,3);
        \draw(11,-0.5) arc (180:165:3);
        \draw (8,1)--(4,2)--(6,10)--(10,9)--cycle;
        \node at(9,0){$\varphi$};
        \filldraw (7,5.5)circle(0.1);
        \filldraw (4,2)circle(0.1);
        \filldraw (6,1.5)circle(0.1);
        \draw[->](6,1.5)--(4,2);
        \draw[->](6,1.5)--(7,5.5);
        \draw[->](7,5.5)--(5,3.5);
        \draw[dashed](12,5.5)--(0,5.5);
        \draw[->](7,5.5)--(7,2.5);
        \draw[dashed](5,3.5)--(5,0.5);
        \draw[dashed](5,0.5)--(7,2.5);
        \draw[->](7,5.5)--(5,0.5);
        \draw[->](6,5.5) arc(180:225:1);
        \node at(5.5,5){$\theta$};
        \draw(6,4.5)--(4,4.5)node[left]{ $\boldsymbol{F}_a=m\boldsymbol{a}$};
        \draw(7,4.2)--(10,4.2)node[right]{ $\boldsymbol{F}_g=m\boldsymbol{g}$};
        \node at(7,6.3){$O$};
        \node at(7.2,1.8){$R$};
        \node at(3.5,1.1){$C$};
        \draw(6,2.8)--(10,2.8)node[right]{$\boldsymbol{F}_{obj}$};
        \draw[->](5.5,1.3)arc(180:360:0.5);
        \node at(6,0.2){$\boldsymbol{\alpha}$};
    \end{tikzpicture}}
    \subfloat[Acceleration phase of the trajectory. \label{fig:alg}]{\includegraphics[width=0.46\textwidth]{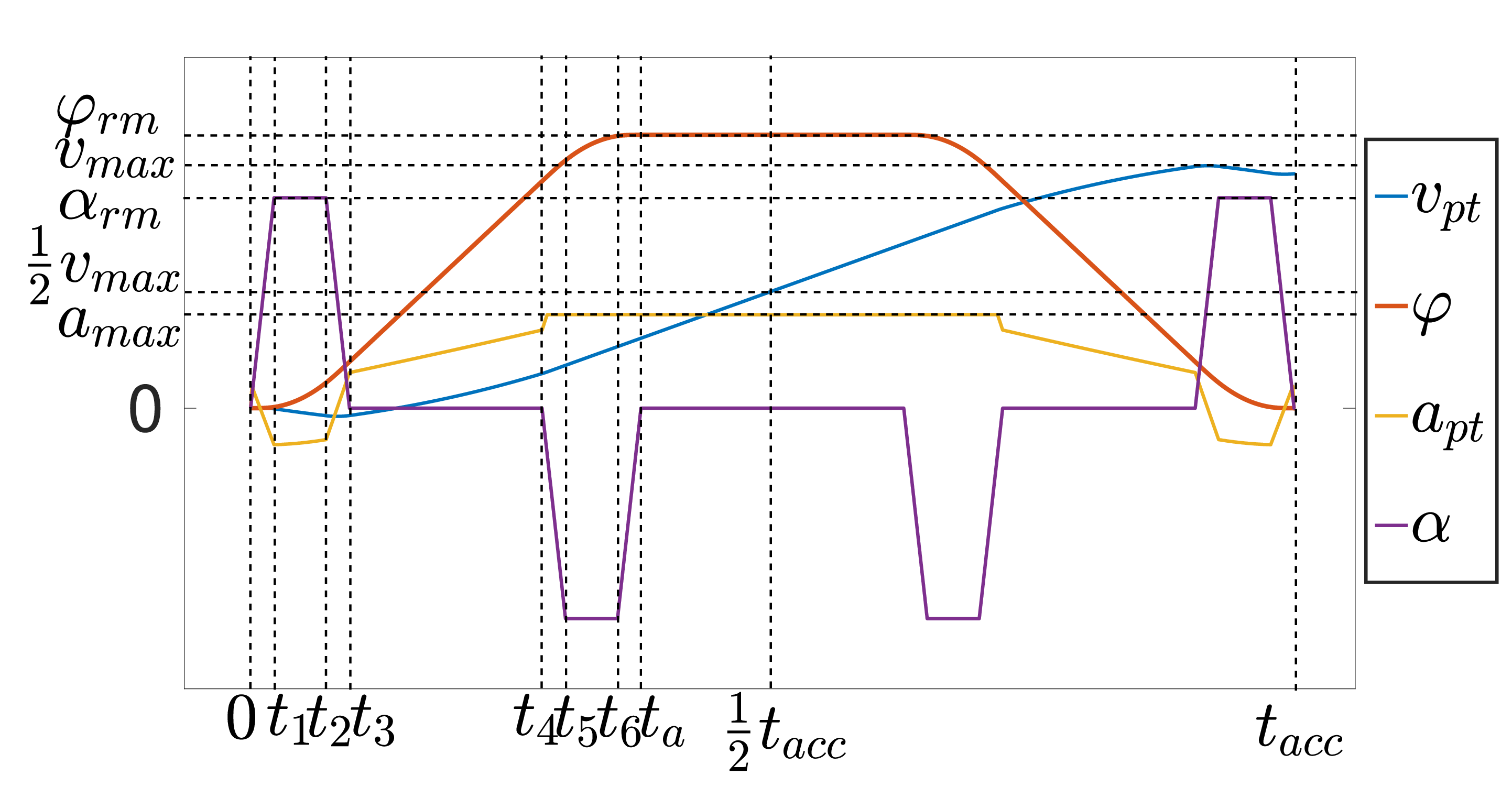}}
    \vspace{-18pt}
    \caption{Physical model and trajectory illustration.} 
    \label{fig:test}
    \vspace{-30pt}
\end{figure}
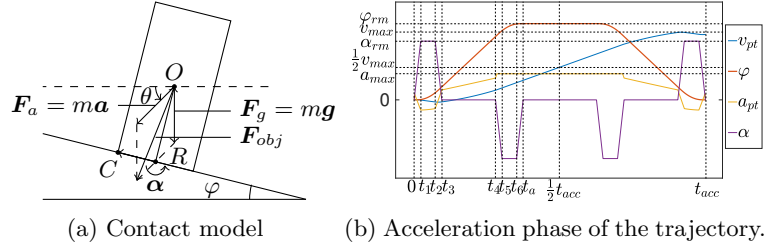

To ensure stability during the transportation process, we aim for no relative motion between the object and the tray. Let the tray rotate around the center of the object's base, and assume  at this point there is a fixed joint between the object and the tray, connecting them. According to the 6D rigid body contact model proposed in \cite{bouchard20156d}, whether the object tilts or not depends on the location of the pressure center. To maximize the acceleration during the transportation process, we assume that the pressure center is located on the boundary of the contact surface between the object and the tray. Based on the rotational motion state of the tray, the torque $\boldsymbol{\tau}=I\boldsymbol{\alpha}$ acting on the virtual fixed joint can be calculated, where $I$ is the moment of inertia, and $\boldsymbol{\alpha}$ is the rotational angular acceleration.  Choosing the tray as the reference frame and analyzing the motion and forces acting on the object as shown in Fig. \ref{fig:phy}, Where $\boldsymbol{a}$ is the object's translational acceleration, $\theta$ is the target direction, $\varphi$ is the current rotation angle of the tray, $O$ represents the object's center of gravity, $R$ is the center of the object's base, and $C$ is the chosen center of pressure position. The object is subjected to a resultant force, denoted as $\boldsymbol{F}_{obj}$, which is a vector sum of the forces due to gravity and acceleration, expressed as $\boldsymbol{F}_{obj} = -m\boldsymbol{a}+m\boldsymbol{g}$. The tray applies a force $\boldsymbol{F}_{tray}$ on the object, which is equal in magnitude but opposite in direction to $\boldsymbol{F}_{obj}$, thus $\boldsymbol{F}_{tray} = -\boldsymbol{F}_{obj}$. Additionally, the object is subjected to a centrifugal force $\boldsymbol{F}_r$, calculated as $\| \boldsymbol{F}_r \| = m\| \boldsymbol{\omega}\|^2\frac{h}{2}$, where $\omega$ represents the rotational velocity, $r$ is the radius of the cylinder, and $h$ denotes the cylinder's height. The object's motion, synchronized with that of the tray and rotating around the object's base center, is facilitated by the collective influence of three forces. This dynamic interaction is encapsulated in the equation below:
\begin{equation}\label{eq1}
    \boldsymbol{\tau} = \overrightarrow{RO} \times \boldsymbol{F}_{obj} + \overrightarrow{RC} \times \boldsymbol{F}_{tray} +  \overrightarrow{RC} \times \boldsymbol{F}_r 
\end{equation}
The objective is to determine the maximum translational acceleration of the object under the current tray motion. Therefore, by rearranging Equ. \ref{eq1}, we can obtain:
\begin{equation}\label{eq2}
     \| \boldsymbol{a} \| =\frac{\frac{I\|\boldsymbol{\alpha}\|}{m}+\frac{h\|\boldsymbol{g}\|}{2}\sin{\varphi}+r\|\boldsymbol{g}\|\cos{\varphi}-\frac{r\|\boldsymbol{\omega}\|^2h}{2}}{\frac{h}{2}\cos{\theta}\cos{\varphi}-\frac{h}{2}\sin{\varphi}\sin{\theta}-r\sin{\varphi}\cos{\theta}-r\cos{\varphi}\sin{\theta}}
\end{equation}
\subsection{Trajectory Planning}\label{IIIB}
In the domain of robotic motion planning, the constraints primarily encompass jerk, acceleration, and velocity parameters. These constraints remain uniform throughout the trajectory, represented as $j \in [-j_{max}, j_{max}]$, $a \in [-a_{max}, a_{max}]$, and $v \in [0, v_{max}]$. Similarly, rotational motion adheres to constraints denoted as $j_r \in [-j_{rm}, j_{rm}]$, $\alpha_r \in [-\alpha_{rm}, \alpha_{rm}]$, and $\omega_r \in [0, \omega_{rm}]$. It's worth noting that robots can achieve high jerks, facilitating rapid acceleration changes. However, since the rotation speed of the tray is comparatively slower, we assume that during the trajectory's acceleration phase, the maximum acceleration an object can reach is primarily dictated by the tray's rotation speed. Therefore, the trajectory planning strategy first generates the rotational motion trajectory of the tray using an S-curve, then calculates the corresponding translational motion trajectory using Equ. \ref{eq2}. The trajectory planning can be summarized in Tab. \ref{tab:rot}.

\begin{table}[b]\tiny
    \centering
    \begin{tabular}{|c|c|c|c|c|c|c|c|c|}
        \hline
            & $t < t_1$ & $t_1 \leq t < t_2$ & $t_2 \leq t < t_3$ & $t_3 \leq t < t_4$ & $t_4 \leq t < t_5$ & $t_5 \leq t < t_6$ & $t_6 \leq t < t_a$ & $t_a \leq t < \frac{1}{2}t_{acc}$\\
        \hline
            $j_r$ & $j_{rm}$ & $0$ & $-j_{rm}$ & $0$ & $-j_{rm}$ & $0$ & $j_{rm}$ & $0$ \\
        \hline
            $\alpha$ & $\uparrow$ & $\alpha_{rm}$ & $\downarrow$ & $0$ & $\downarrow$ & $-\alpha_{rm}$ & $\uparrow$ & $0$\\
        \hline
            $\omega$ & $\uparrow$ & $\uparrow$ & $\uparrow$ & $\omega_{rm}$ & $\downarrow$ & $\downarrow$ & $\downarrow$ & $0$\\
        \hline
            $\varphi$ & $\uparrow$ & $\uparrow$ & $\uparrow$ & $\uparrow$ & $\uparrow$ & $\uparrow$ & $\uparrow$ & $\varphi_{rm}$\\
        \hline
            $a$ & \multicolumn{7}{|c|}{Calculated based on(\ref{eq2})} & $a_{max}$\\
        \hline
            $v$ & \multicolumn{8}{|c|}{Increasing and reaching $\frac{1}{2}v_{max}$ at time $\frac{1}{2}t_{acc}$.}\\
        \hline
    \end{tabular}
    \caption{Changes in motion states during the trajectory acceleration phase.}
    \label{tab:rot}
    \vspace{-10pt}
\end{table}

The next step is to ensure that the trajectory does not exceed the robot's motion constraints. It can be observed that at time $t_a$, the trajectory's acceleration reaches its maximum, while the velocity is half of the maximum speed. Consequently, we can formulate the following conditions:
\begin{equation}
    a_{pt}(t_a) \leq a_{max},
\end{equation}
\begin{equation}
    v_{pt}(t_a) \leq \frac{v_{max}}{2}.
\end{equation}

Finally, assuming there is no constant velocity phase in the trajectory, the velocity at time $t_a$ precisely equals the average velocity of the trajectory. By knowing the durations of the acceleration phases $t_a$ and deceleration phases $t_b$, the following conditions ensure that the trajectory does not exceed the target point. 
\begin{equation}
    2v_{pt}(t_a)(t_a+t_b) \leq p_t.
\end{equation}
Adjusting the duration of the constant velocity phase based on the remaining distance between the trajectory and the target point completes the trajectory planning.
\vspace{-10pt}
\section{Experiment and Results}
\begin{figure}[h]
    \centering
    \vspace{-25pt}
    \subfloat[Experiment setup \label{fig:set}]{\includegraphics[width=0.40\textwidth]{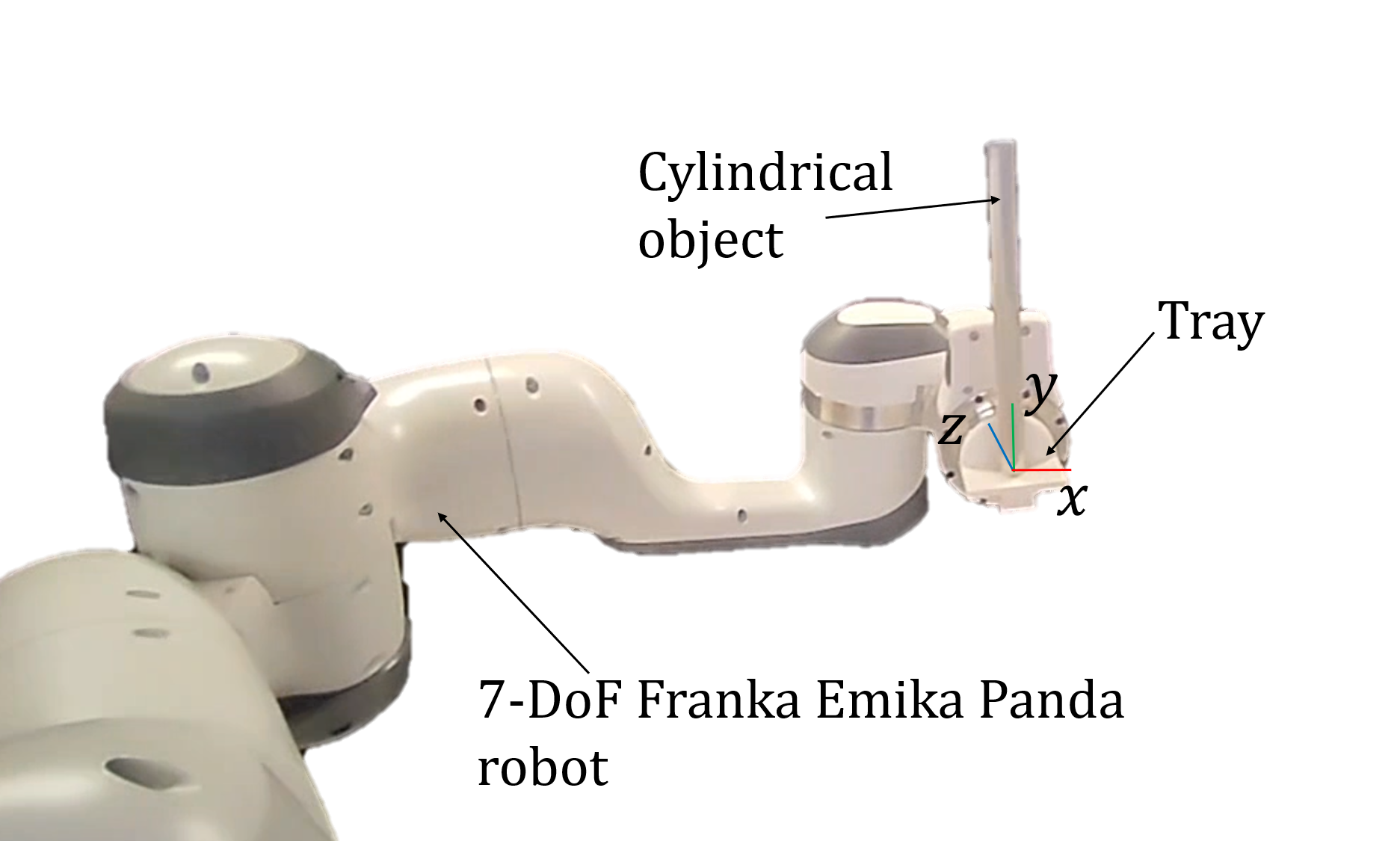}}
    \subfloat[Theoretical efficiency \label{fig:theo}]{\includegraphics[width=0.40\textwidth]{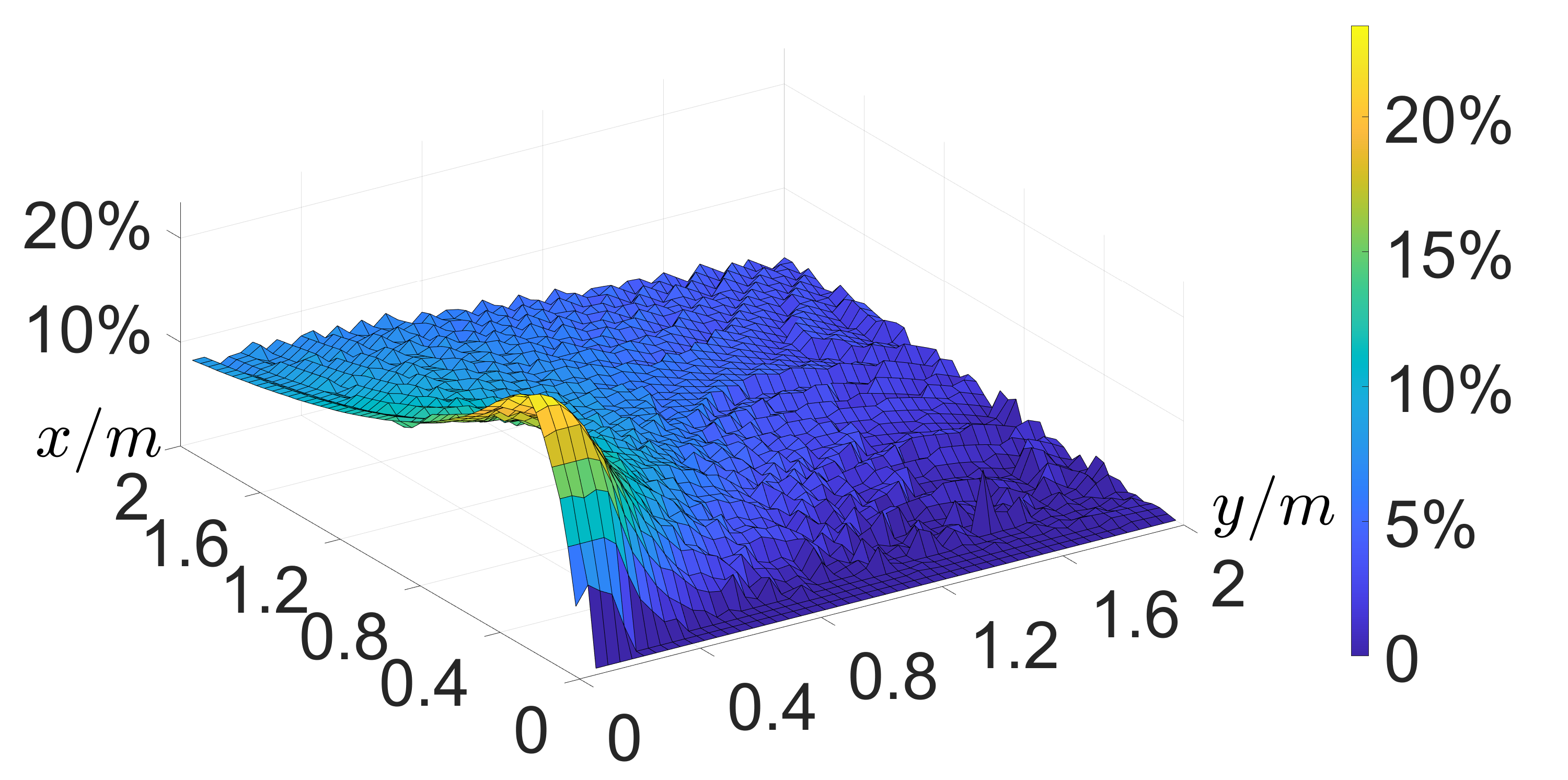}}
    \\
    \vspace{-10pt}
    \subfloat[The target and actual trajectory \label{fig:follow}]{\includegraphics[width=0.40\textwidth]{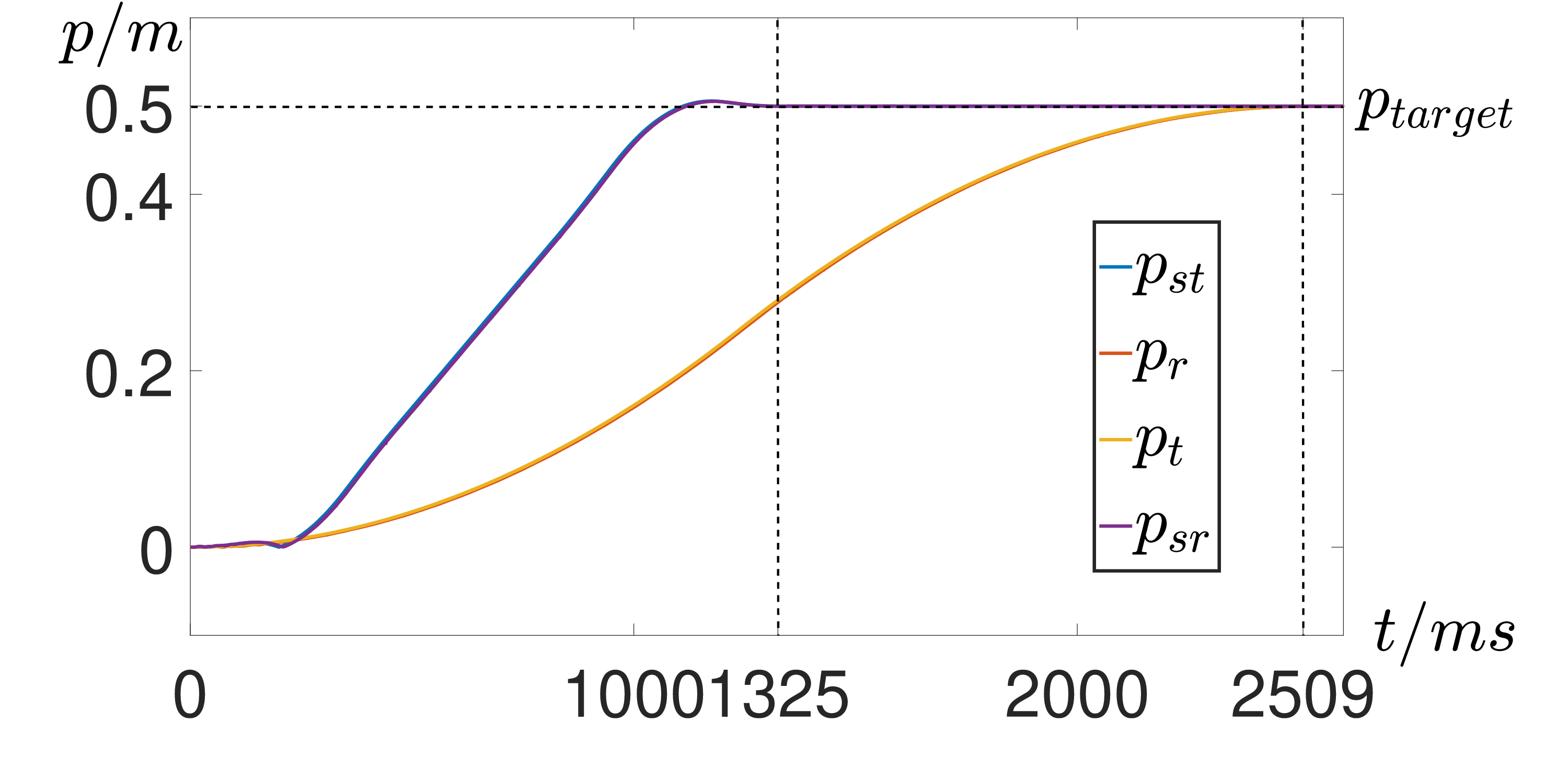}}
    \subfloat[End-effector rotation \label{fig:nor}]{\includegraphics[width=0.40\textwidth]{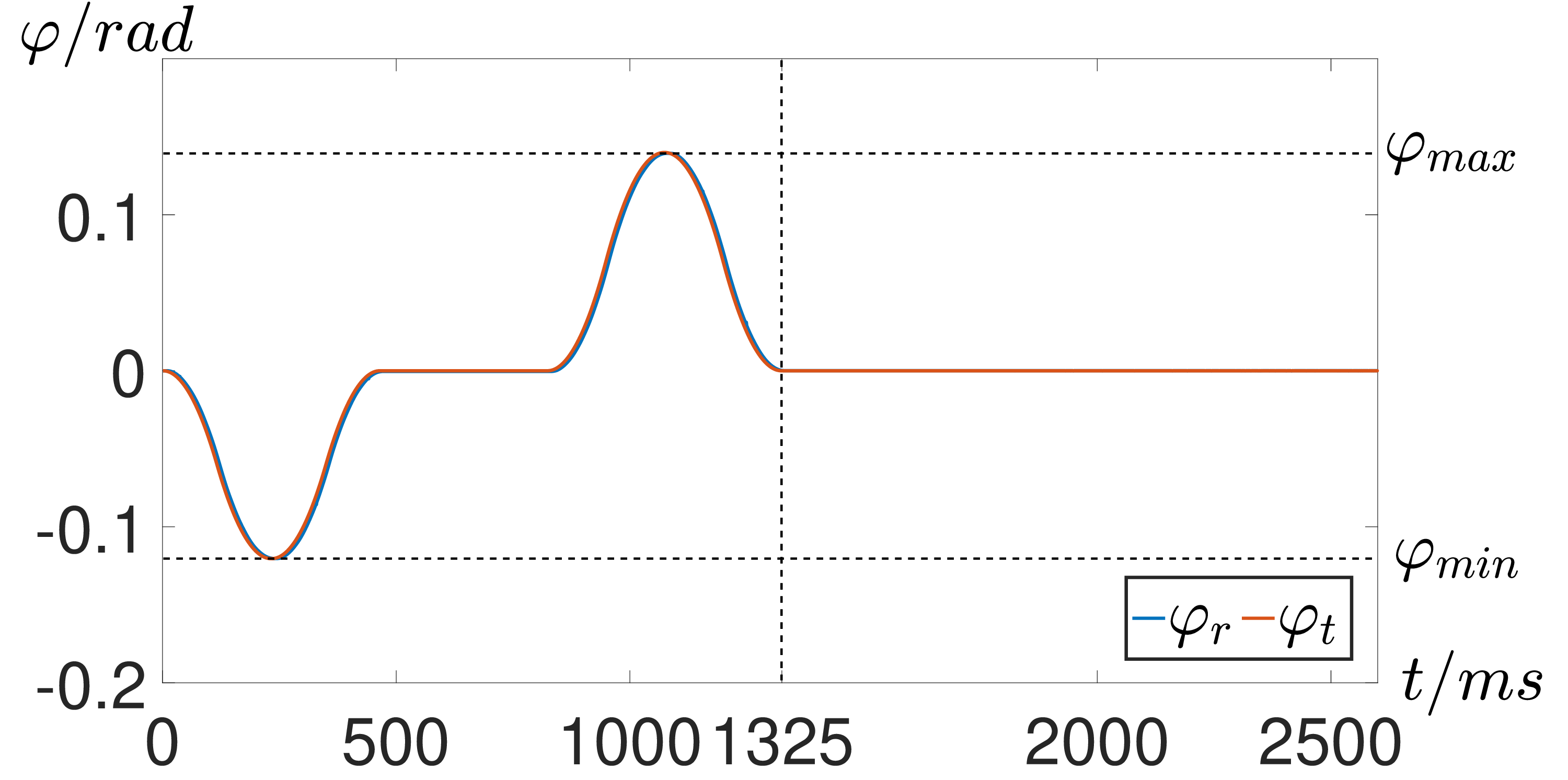}}
    \vspace{-10pt}
    \caption{Experiment results.} 
    \vspace{-20pt}
    \label{fig:test}
\end{figure}

To test the effectiveness and time efficiency of our method, we conducted a series of experiments using a 7-DoF Franka Emika Panda robot. The setup for these experiments is depicted in Fig. \ref{fig:set}. The following conditions are selected as motion constraints in the experiment: $j_{max}=6500\ m/s^3$, $a_{max}=13\ m/s^2$,$v_{max}=0.6\ m/s$, $j_{rm}=6000\ rad/s^3$, $a_{rm}=9\ rad/s^2$,$ v_{rm}=2.61\ rad/s$. The tested object is a uniform aluminum cylinder with a radius of $r=8\ mm$ and a height of $h=0.2\ m$. We performed theoretical calculations based on these conditions and compared the time taken by our proposed method with the time taken by the method that does not involve tray rotation. The results are shown in Fig. \ref{fig:theo}, where $x$ represents the horizontal displacement of the target, and $y$ represents the vertical displacement of the target. Our proposed method can improve time efficiency by up to $25\%$. 
\par 
The experimental design sets the target distance to $p=0.5\ m$, with the target located above the starting point and orientation angle $\theta=\frac{\pi}{8}$. Due to the assumption in the physical model and control errors of the robot, the experiment gradually reduces the radius of the object's base until a stable trajectory is found. The time required for the trajectory is then compared with the time efficiency of the trajectory planned using S-curves under the same motion constraints.

In the experiments\footnote{Videos shown in the \href{https://www.youtube.com/playlist?list=PLpV4tWGLzYVVn58opWrHWbIj-QrLsyJ7j}{link}.}, the input radius for stable trajectories obtained are $r_o=3\ mm$ without rotation and $r_r=4\ mm$ with rotation. The trajectories of the target and the robot's motion are shown in Fig. \ref{fig:follow} and Fig. \ref{fig:nor}. 
It can be seen that our proposed new method reduces the time taken by $47.2\%$ compared to the motion trajectory without rotation.
\vspace{-12pt}
\section{Conclusion}
\vspace{-5pt}

This study introduces a novel trajectory planning approach, integrating tray rotation, for the non-prehensile transportation of unstable objects. Our method significantly accelerates straight-line transportation in three-dimensional space, and its efficacy has been demonstrated through practical implementation on a real robot. In future work, our aim is to enhance the proposed method to enable trajectory planning for the non-prehensile transportation of unstable objects along arbitrary paths.

\vspace{0.5em}
%
%
\noindent \small \textbf{Acknowledgement.} The work was funded by the German Research Foundation (DFG, Deutsche Forschungsgemeinschaft) as part of Germany’s Excellence Strategy – EXC 2050/1 – Project ID 390696704 – Cluster of Excellence “Centre for Tactile Internet with Human-in-the-Loop” (CeTI) of Technische Universität Dresden. The work was also supported by the European Union’s Horizon 2020 research and innovation programme as part of the project euROBIN under grant no. 101070596.

\end{document}